\documentclass[letterpaper, 10 pt, conference]{ieeeconf}  

\IEEEoverridecommandlockouts                              

\usepackage{graphics} 
\usepackage{epsfig} 
\usepackage{mathptmx} 
\usepackage{times} 
\usepackage{amsmath} 
\usepackage{amssymb}  
\usepackage{pifont}   
\usepackage[absolute,overlay]{textpos}
\usepackage{multirow}
\usepackage{multicol}
\usepackage{graphicx}
\usepackage{amsmath}
\usepackage{amssymb}
\usepackage{booktabs}
\usepackage{xcolor,colortbl}
\usepackage{makecell}
\usepackage{pifont}
\usepackage{balance}
\usepackage{cancel}
\usepackage{cite}
\usepackage{adjustbox}
\usepackage{soul}
\usepackage{tabularx}
\usepackage{pifont}
\usepackage{amssymb}

\newcommand{\xiaoyang}[1]{{\color{black}#1}}

\newcommand{\yubin}[1]{{\color{black} #1}}

\newcommand{\shortname}[0]{{\textit{SynthDrive}}}
\newcommand{\eg}{\textit{e.g.}}

\title{\LARGE \bf
\shortname: Scalable \textit{Real2Sim2Real} Sensor Simulation Pipeline for High-Fidelity Asset Generation and Driving Data Synthesis
}

\author{Zhengqing Chen$^{1,*}$, Ruohong Mei$^{1,*}$, Xiaoyang Guo$^{1,\dag}$, Qingjie Wang$^{1}$, Yubin Hu$^{2}$,\\Wei Yin$^{1}$, Weiqiang Ren$^{1}$ and Qian Zhang$^{1}$  
\thanks{$^{1}$ Horizon Robotics. $^{2}$ Tsinghua University}
\thanks{$^{*}$ Equal contribution $\dag$ Corresponding Author.}%
}

\begin{document}

\maketitle
\thispagestyle{empty}
\pagestyle{empty}


\begin{figure*}[!htp]
    \centering
    \includegraphics[width=1.0\textwidth]{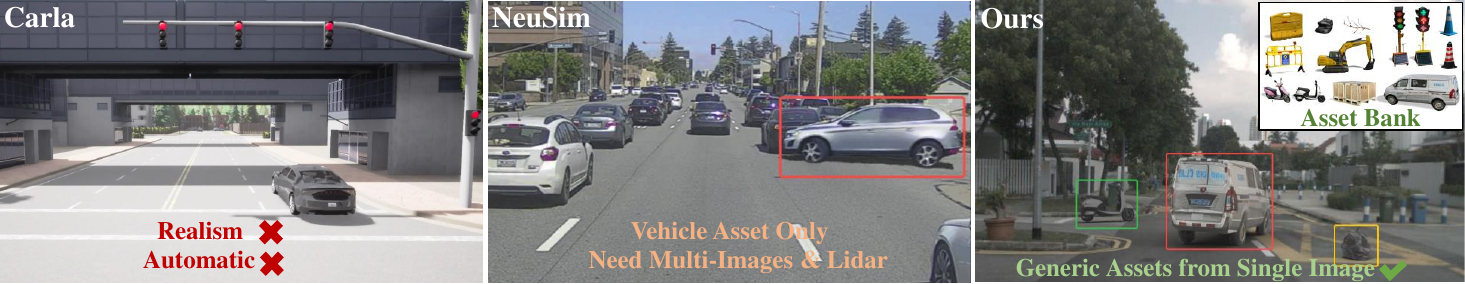}
        \vspace{-7mm}
    \caption{Comparison of different approaches for autonomous driving simulation. Left: CARLA\cite{dosovitskiy2017carla}, a CG-based platform with manually crafted assets, lacking realism and scalability. Middle: NeuSim\cite{yang2023neusim}, a vehicle-only asset reconstruction approach requiring complex multi-modality images and LiDAR data. Right: Our approach, a fully automated data synthesis framework capable of generating diverse generic assets. The figure shows the automatically-constructed asset bank that includes traffic elements, vehicles, and road infrastructure.}
    \label{fig:teaser}
    \vspace{-7mm}
\end{figure*}

\begin{abstract}



In the field of autonomous driving, sensor simulation is essential for generating rare and diverse scenarios that are difficult to capture in real-world environments.
Current solutions fall into two categories: 1) CG-based methods, such as CARLA, which lack diversity and struggle to scale to the vast array of rare cases required for robust perception training; and 2) learning-based approaches, such as NeuSim, which are limited to specific object categories (vehicles) and require extensive multi-sensor data, hindering their applicability to generic objects.
To address these limitations, we propose \shortname, a scalable \textit{``real2sim2real''} system that leverages 3D generation to automate asset mining, generation, and rare-case data synthesis.

Our framework introduces two key innovations: 1) \textit{Automated Rare-Case Mining and Synthesis.} Given a text prompt describing specific objects, {\shortname} automatically mines image data from the Internet and then generates corresponding high-fidelity 3D assets, which eliminates the need for costly manual data collection. By integrating these assets into existing street-view data, our pipeline produces photorealistic rare-case data, supporting rapid scaling to diverse assets including irregular obstacles and temporary traffic facilities. 2) \textit{High-Fidelity 3D Generation.} We propose a hybrid asset generation pipeline that combines a geometry-aware LRM, iterative mesh optimization, and an improved texture fusion algorithm. Our approach achieves 0.0164 Chamfer Distance on the GSO dataset, outperforming InstantMesh by 14.1\% in geometry accuracy, and achieves 19.05 PSNR (vs 16.84) for texture quality. This enables fine geometry details and high-resolution texture generation, which is essential for perception model training. 
Experiments demonstrate that \shortname-generated data improves the performance of downstream perception tasks (2D and 3D detection on
rare objects) by 2-4\% mAP. 
{\shortname} greatly lowers the data production cost and improves the diversity for corner-case data generation, showcasing extensive potential applications in the field of autonomous driving.

\end{abstract}

\section{INTRODUCTION}

The development of robust autonomous driving systems underscores the critical need for the coverage and diversity of training data and rare scenarios, particularly those involving safety-critical corner cases.
However, relying solely on collecting data from the real world is insufficient in capturing all rare objects and corner cases~\cite{wang2021advsim}. Therefore, simulating these infrequent traffic scenarios is paramount for the development of robust and secure vehicles.

Existing computer graphics-based software, such as Carla~\cite{dosovitskiy2017carla}, use graphic rendering engines to depict these corner cases. However, they heavily rely on manually crafted 3D assets to construct specific driving scenes. 3D assets in such systems typically encompass a limited range of textures and lack realism, which restricts their capability to fully replicate the complexity and diversity of real-world environments.
Another line of research works~\cite{zhang2021ners,yang2023neusim,wang2023cadsim,chen2021geosim} employ 3D reconstruction techniques instead of handcrafting to create 3D assets, which greatly improves the realism of 3D assets. The reconstructed 3D models are then inserted into existing driving scenes to create rare scenarios. 
To further improve realism, lighting estimation~\cite{wang2023fegr,wang2022neurallightfield,pun2023lightsim} and style transfer~\cite{bai2024anythinginanyscene} can be additionally applied to ensure the correctness of environment lighting and realism. However, these methods investigate 3D reconstruction only for limited object categories like vehicles and require complex multi-sensor inputs, lacking flexibility and scalability to generic objects. 
Different from previous methods, recent image-to-3D methods ~\cite{poole2022dreamfusion,tochilkin2024triposr,xu2024grm,xu2024instantmesh,liu2023zero123,shi2023zero123++,liu2024one2345,liu2024one2345++,wu2024unique3d} could generate diverse 3D assets from single input images, offering the potential to address the above issues in current data synthesis approaches.

\begin{figure*}[t]
    \centering
    \includegraphics[width=1.0\textwidth]{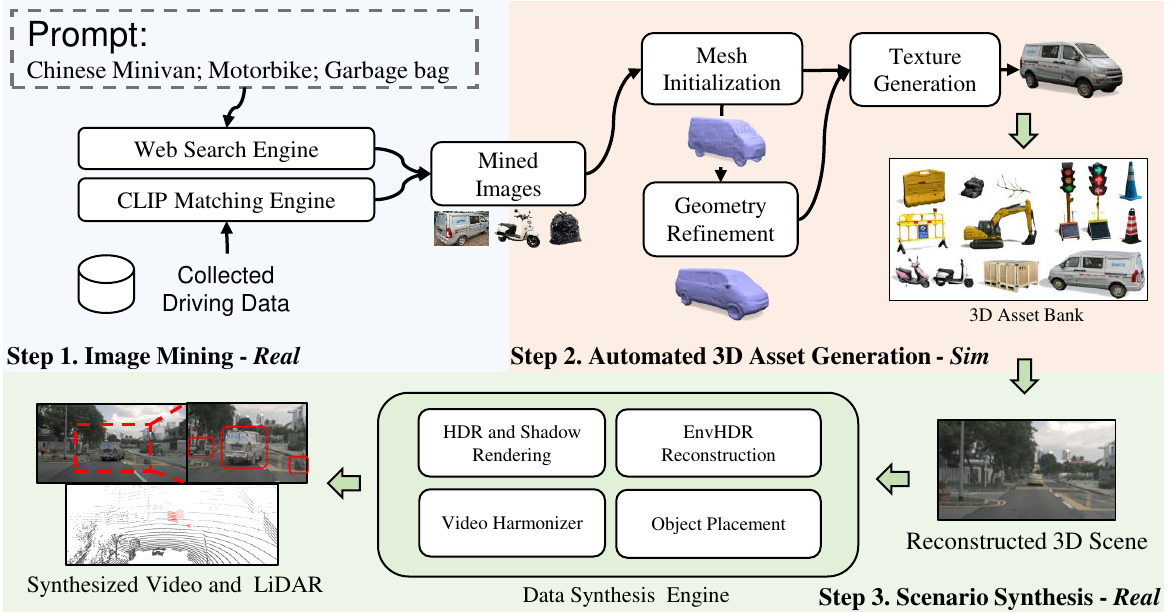}
    \caption{The Framework of {\shortname}. 1) Given a text \textit{prompt}, our system automatically mines asset images from existing driving data via CLIP~\cite{radford2021clip}, or from web search engines. 2) High-quality 3D assets are reconstructed from these images using the proposed hybrid multi-stage image-to-3D algorithm, producing a comprehensive 3D asset bank. 3) Synthetic data is finally rendered by blending real images (or reconstructed 3D scenes~\cite{mildenhall2021nerf,wang2021neus,kerbl20233dgaussian}) with 3D assets queried from the asset bank in a large-scale scalable way.}
    \label{fig:pipeline}
        \vspace{-5mm}
\end{figure*}

Inspired by the above observations, we introduce {\shortname} (Fig.~\ref{fig:pipeline}), a \textit{``real2sim2real''} data generation pipeline to overcome the realism and scalability issues in existing pipelines~\cite{dosovitskiy2017carla,yang2023neusim,wang2023cadsim,chen2021geosim}. The core concept of this approach is \textit{fully-automated} rare-data mining and synthesis.
Our objectives address four critical gaps in existing methods:
1) \textbf{Generic Object Reconstruction}: Unlike NeuSim~\cite{yang2023neusim} and CADSim~\cite{wang2023cadsim}, which are confined to limited object categories, our system supports arbitrary objects (e.g., portable traffic lights, fallen debris) through automated asset mining.
2) \textbf{Low Data \& Computation Cost}: In contrast to [7] requiring complex sensor inputs, \shortname reconstructs assets from a single image within 1 minute, without the need for additional manual data collection.
3) \textbf{High-Fidelity Geometry \& Texture}: the proposed hybrid image-to-3D pipeline incorporates robust geometric and texture details, which is more suitable for perception model training compared with~\cite{xu2024instantmesh,long2024wonder3d}.

To achieve these objectives, {\shortname} is built with three core modules: \textbf{Asset Mining, 3D Asset generation, and Scenario Synthesis}. Given a text prompt describing specific objects (\eg, ``portable traffic lights" or ``irregular obstacles"), our system automatically mines image data from existing driving data and web-collected sources using CLIP-guided~\cite{radford2021clip} retrieval. This eliminates the need for manual data collection and ensures the coverage of diverse rare objects for simulation.
The key of generic 3D asset reconstruction lies in our high-fidelity 3D asset generation module, which transforms single images into detailed 3D models. We propose a hybrid image-to-3D pipeline that combines a geometry-aware LRM, iterative mesh
optimization, and an improved texture fusion algorithm. 
Specifically, starting with a single input image, we first extract high-quality texture and geometry cues using a multi-view diffusion model~\cite{shi2023zero123++} and an image-to-normal diffusion model~\cite{ye2024stablenormal}.
\yubin{The multi-view texture cues are utilized to initialize a coarse 3D model with a feed-forward large reconstruction model~\cite{xu2024instantmesh},}
\yubin{and the geometric surface normal cues are utilized to refine the coarse model through a normal-supervised explicit mesh optimization process, enabled by differentiable surface normal rendering.}
To further improve texture quality, we up-sample the texture cues by diffusion-based image super-resolution~\cite{arkhipkin2023kandinsky}. \yubin{The high-resolution images are then comprehensively mapped and fused } across multiple views to produce the final textured 3D models, maintaining high-fidelity details.

Finally, \yubin{the} generated 3D assets are integrated into existing driving videos or reconstructed simulation environments for further synthetic data generation and closed-loop simulations. A complex HDR lighting estimation and post-video harmonizer are employed to ensure the correctness of shadows and colors. The road surface geometry is also reconstructed to ensure the correct placement of inserted 3D assets. It offers enhanced realism and diversity for the simulated scenarios, which are crucial for the rapid development of autonomous driving systems.
To sum up, our contributions are three-fold: 
\begin{itemize}
    \item We introduce \shortname, a cost-effective and scalable \textit{``real2sim2real''} pipeline for automatic asset library construction and data synthesis for autonomous driving. 
    \item We propose a novel single-view reconstruction approach with enhanced mesh geometry optimization and detailed texture fusion to create high-fidelity and high-resolution generic 3D assets. 
    \item We conduct extensive experiments on public datasets. Results demonstrate that our 3D asset generation approach surpasses state-of-the-art approaches in terms of geometric accuracy, rendering quality, view generalization, and versatility. Our synthesis data significantly improves the performance of downstream perception tasks(2D and 3D detection on rare objects).
\end{itemize}



\section{Related Work}
\subsection{Asset Reconstruction}
In the domain of autonomous driving simulations, various methodologies have been explored for 3D asset reconstruction and generation. As illustrated in Fig.~\ref{fig:teaser}, 
Computer Graphics (CG)-based platforms like CARLA~\cite{dosovitskiy2017carla} offer comprehensive simulation tools but rely heavily on manually crafted 3D assets, which limits its scalability and adaptability.
Techniques based on neural radiance fields (NeRFs)~\cite{mildenhall2021nerf,muller2022instantngp,tancik2022blocknerf,barron2022mipnerf360,barron2023zipnerf}, NeuS~\cite{wang2021neus,li2023neuralangelo,yu2022monosdf,yang2023unisim,yariv2023bakedsdf} and recent 3D Gaussian Splatting~\cite{kerbl20233dgaussian,yu2024mipsplatting,zhou2024drivinggaussian,yan2024streetgaussian} require high-quality 360-view scans of objects. However, objects in autonomous driving data are usually sparsely observed, leading to suboptimal geometry and texture quality in occluded areas.

To address the above issues, several approaches~\cite{chen2021geosim,wang2023cadsim} incorporate additional priors to enhance vehicle reconstruction quality. 
GeoSim~\cite{chen2021geosim} incorporates multi-view supervisions and LiDAR-depth observations into a mesh reconstruction network for 3D asset generation.
CADSim~\cite{wang2023cadsim} combines part-aware CAD models with differentiable rendering to reconstruct vehicle geometry. NeuSim~\cite{yang2023neusim} utilizes a neural volume rendering approach to reconstruct vehicles from in-the-wild images and LiDAR data. 
However, these methods primarily focus on the reconstruction of vehicle assets, relying on extensive multi-view and multi-sensor data, which limits their applicability to specific objects like cars. In contrast, our method requires only a single image to reconstruct the object with significantly finer details. Concurrently, research works like 3DRealCar~\cite{du20243drealcar} and DreamCar~\cite{du2024dreamcar} collect 360-degree car datasets and employ diffusion priors for 3D asset generation. However, their focus remains solely on vehicles, thus limiting their broader applications.


To synthesize corner-case data, the reconstructed 3D assets are inserted into existing driving scenes for scene editing. To enhance realism, LightSim~\cite{pun2023lightsim} proposes to synthesize data under different lightings. \cite{wang2022neurallightfield} recovers a 5D HDR light field for relighting inserted objects. FEGR~\cite{wang2023fegr} employs an inverse rendering technique to recover the lighting conditions of existing images, enhancing the virtual object insertion realism. However, these methods did not address the scalability issues for asset library construction.



\subsection{Single View 3D Reconstruction}
Singe view 3D reconstruction~\cite{poole2022dreamfusion} aims to reconstruct objects from only a single image as input. One line of work~\cite{poole2022dreamfusion,qian2023magic123,tang2023makeit3d,wang2023scorejacobian} utilizes diffusion generative priors from pretrained image diffusion models~\cite{rombach2022stablediffusion,podell2023sdxl}, guided by score distillation sampling loss (SDS)~\cite{poole2022dreamfusion}. However, these approaches usually suffer from ``multi-face problems'' and time-consuming per-scene optimization. To accelerate the generation, Zero123~\cite{liu2023zero123} demonstrates that diffusion models~\cite{rombach2022stablediffusion} can be fine-tuned to generate novel view images from a single-view input conditioned on target camera poses. Subsequent works~\cite{shi2023zero123++,liu2023syncdreamer,shi2023mvdream} further improve the consistency of multi-view predictions by simultaneously generating multi-views of an object. 3D inconsistency from those generated multi-view images can impact the geometric quality of the 3D models. Wonder3D~\cite{long2024wonder3d} proposes to predict multi-view normals as supplementary signals and adapt SDF-based 3D reconstruction techniques to ensure robustness. One2345~\cite{liu2024one2345,liu2024one2345++} directly regresses plausible 3D shapes from the images generated from Zero123~\cite{liu2023zero123}.

More recently, large reconstruction models~\cite{hong2023lrm,tochilkin2024triposr,xu2024instantmesh,xu2024grm,wang2024crm} have demonstrated that 3D generation is feasible using feed-forward networks. Large transformer models~\cite{vaswani2017transformer} are trained \yubin{on large-scale 3D datasets~\cite{deitke2023objaverse,deitke2024objaversexl}} to predict diverse 3D representations, such as NeRF~\cite{hong2023lrm}, 3D Gaussian~\cite{xu2024grm}, and mesh~\cite{wang2024crm, xu2024instantmesh}.
Although the above methods achieve significant efficiency, the geometric and texture quality remains suboptimal for asset generation in self-driving simulations. It is crucial to ensure asset details to bridge domain gaps, therefore, in this paper, we extensively explore to improve the geometric and textural details in our proposed framework.



\begin{figure}[t!]
    \centering
    \includegraphics[width=0.50\textwidth]{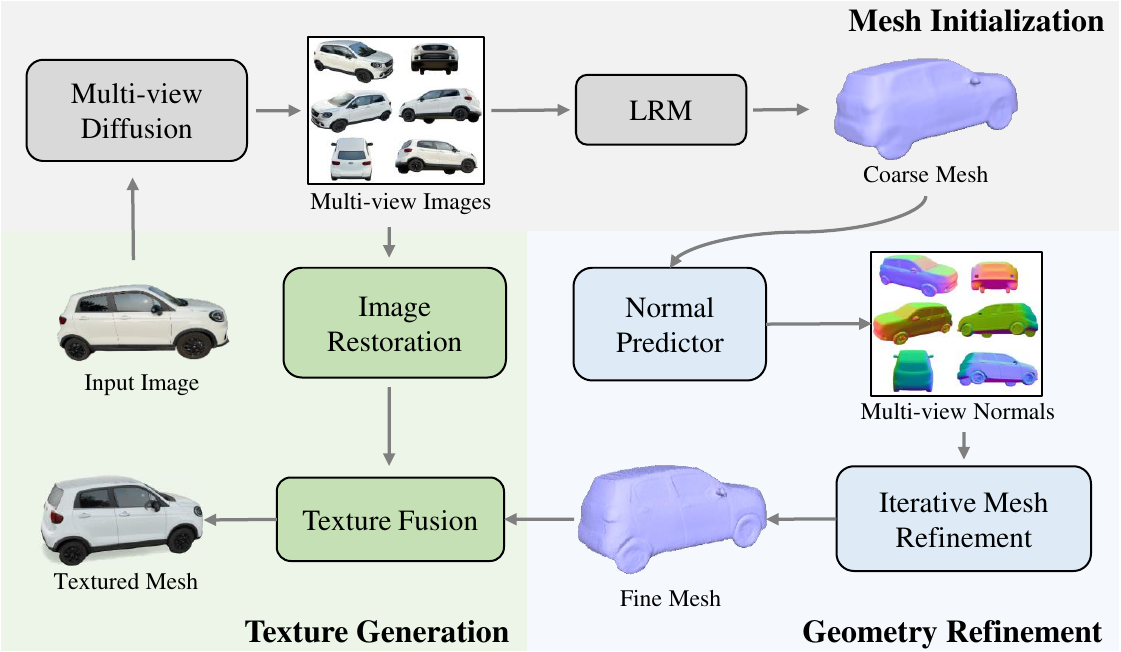}
    \caption{Illustration of the Proposed 3D Asset Generation Pipeline. Given a single input image, multi-view images and normal maps are generated using a multi-view diffusion model~\cite{shi2023zero123++} and a normal estimation model~\cite{ye2024stablenormal}. \xiaoyang{Starting with a coarse mesh reconstructed from LRM, the geometry is refined by iterative normal-aware mesh optimization}. The final texture is enhanced with an improved texture fusion strategy to preserve high-resolution details.}
    \label{fig:method1}
        \vspace{-5mm}
\end{figure}

\begin{figure*}[t]
    \centering
    \begin{tabularx}
    {0.95\linewidth}{c *8{>{\centering}X}} 
    \hspace*{-0.2cm}Ours &\hspace*{1cm}CRM  & \hspace*{0.5cm}InsMesh   & \hspace*{0.3cm}Wonder3D   & \hspace*{0.7cm}Ours &\hspace*{0.7cm}CRM  & \hspace*{0.5cm}InsMesh   & \hspace*{0.3cm}Wonder3D
    \end{tabularx}
    \includegraphics[width=1\textwidth]{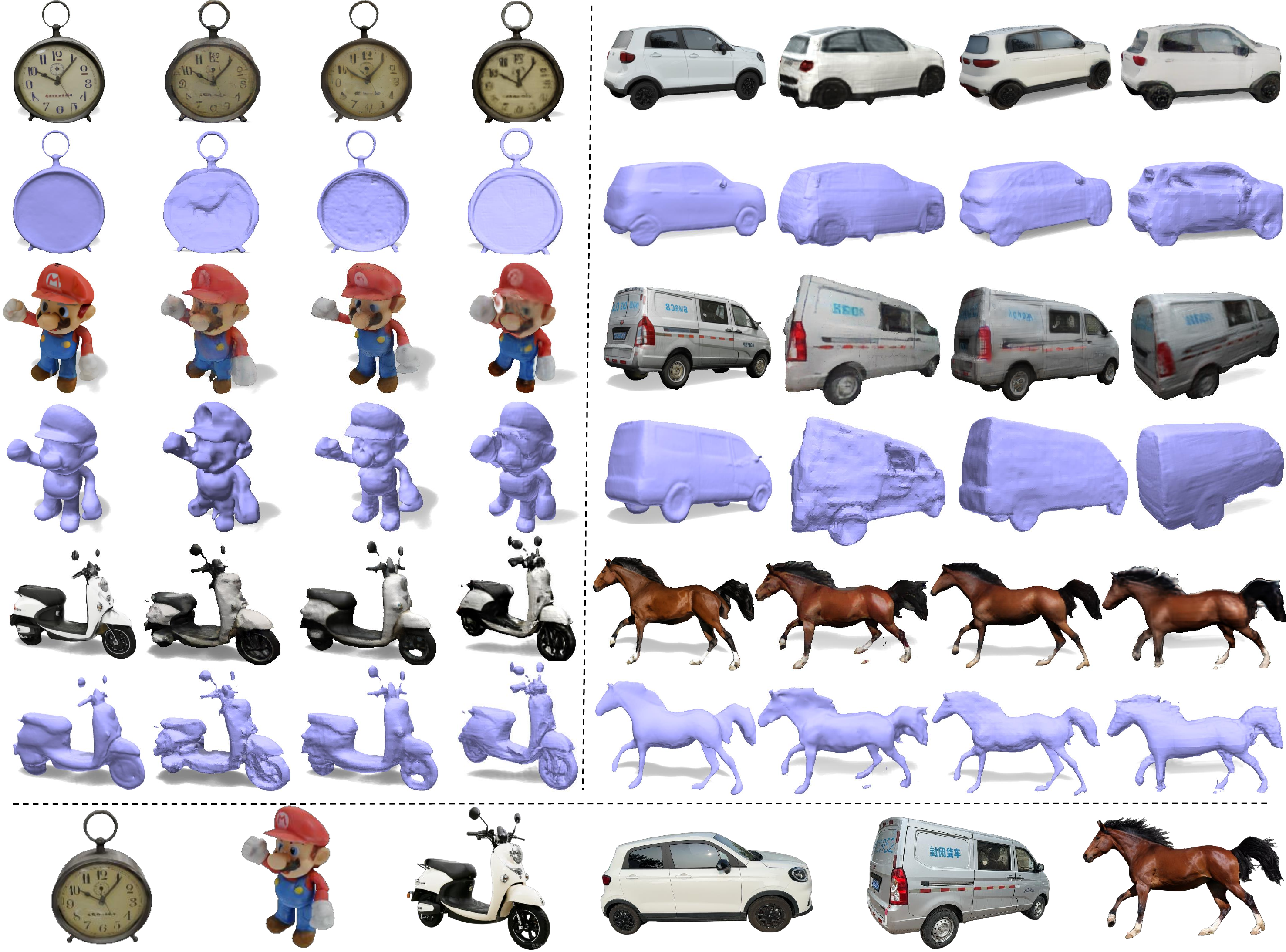}
    \begin{tabularx}
    {0.95\linewidth}{c *6{>{\centering}X}} 
    \hspace*{0.6cm}(a) &\hspace*{1.9cm}(b)  & \hspace*{1.4cm}(c)   & \hspace*{1.15cm}(d)   & \hspace*{1.8cm}(e) & \hspace*{1.6cm}(f)
    \end{tabularx}
    \caption{Qualitative Comparison of Different Methods. (a)-(f) are inputs. Our mesh shows superior geometry and texture quality. Zoom in for more details.}
    \vspace{-7mm}
    \label{fig:results}
\end{figure*}





\section{Methodology}
We propose \shortname, an automated data generation framework tailored for rare data synthesis and autonomous driving simulation. As illustrated in Fig.~\ref{fig:pipeline}, the framework consists of three core modules: \textbf{image mining, automated 3D asset generation, and multi-modal sensor simulations}. The framework begins with CLIP-guided retrieval to mine diverse image data from multiple sources (Sec.~\ref{sec:image-mining}). Next, a hybrid image-to-3D pipeline combines iterative mesh refinement and advanced texturing mapping to generate high-fidelity 3D assets from single-view images (Sec.~\ref{sec:asset-generation}). Finally, the assets are integrated into driving scenes through relighting and video harmonization, combined with 3D reconstruction for accurate object placement (Sec.~\ref{sec:data-synthesis}). 

\subsection{Image Mining}
\label{sec:image-mining}


The pipeline begins with a sophisticated image-mining process. This phase is crucial for reconstructing rare object categories that are essential for robust autonomous driving systems. Given existing driving data, CLIP embeddings~\cite{radford2021clip} of images are pre-computed to facilitate \textit{text-to-image} retrieval.
Text \textit{prompts} (\eg, ``portable traffic lights'' and ``irregular obstacles'') are used to guide the retrieval of images for rare objects. Additionally, images are sourced from web search engines to ensure broader diversity and coverage of image results, complementing the existing database with a wider complexity and variety of challenging scenarios.

\subsection{Automatic 3D Asset Generation from Mined Images}
\label{sec:asset-generation}
\textbf{Mesh Initialization.} 
Generating multi-view images from a single image is fundamental to single-image 3D reconstruction. We employ Zero123++\cite{shi2023zero123++} for multi-view image generation due to its ability to model multiple fixed viewpoints simultaneously, enhancing multi-view consistency and yielding high-quality multi-view images. These images are then used to initialize a coarse 3D model via InstantMesh~\cite{xu2024instantmesh}, a transformer-based large reconstruction model (LRM). While LRM networks~\cite{hong2023lrm,xu2024grm,xu2024instantmesh} generally produce meshes with good topology, they are limited by spatial resolution and lack of geometry details, leading to suboptimal reconstruction precision (see Fig.~\ref{fig:method1} and Fig.~\ref{fig:results}). To address this, we use StableNormal~\cite{ye2024stablenormal} to predict normal maps from multi-view images as geometric cues for subsequent mesh optimization.

\textbf{Iterative Mesh Refinement.} Inspired by prior works~\cite{long2024wonder3d,lu2024direct25d,palfinger2022continuous}, we apply a mesh-based optimization that refines the initial coarse mesh using differentiable rendering. Given the initial coarse mesh, it first undergoes differentiable rendering\cite{laine2020modular} to generate normal maps and silhouette maps, which are used to compute loss with multi-view observations. The vertex positions are updated based on the gradients, followed by explicit surface remeshing, which iteratively applies edge-split, edge-collapse, smoothing, and relocating \cite{palfinger2022continuous}. During each iteration, the initial mesh is subdivided, optimized, and smoothed, enabling the expression of finer geometric details and steering optimization away from local minima.
The loss function of iterative mesh refinement consists of three components:
\[L = L_n + \lambda_{mask}L_{mask} + \lambda_{lap}L_{lap},\]
where ${L}_{{n}} = \sum\limits_{i} \left\| \hat{N}_i - N_i^{pred} \right\|_2^2$ is the L2 loss of normal maps from different views, ${L}_{\text{mask}} = \sum\limits_{i} \left\| \hat{M}_i - M_i^{pred} \right\|_2^2$ represents the alpha mask loss between the rendered images and the silhouette maps. Additionally, we introduce a Laplacian smoothness\cite{sorkine2004laplacian} term \( L_{lap} \) to constrain the consistency of surface normals, resulting in a smoother mesh.

\textbf{Texture Generation.} Realistic appearance is crucial for autonomous driving simulation assets, as it enhances the realism of simulation scenes and reduces the domain gap between synthetic and real data. To address the resolution limitations of multi-view images generated by diffusion models, we adopt a diffusion-based image super-resolution model\cite{arkhipkin2023kandinsky} to upsample the images by a factor of two to recover more texture details. 

Unlike traditional methods that rely on the initial colors provided by LRM networks, we develop an explicit texture fusion algorithm to faithfully preserve texture details from the upsampled multi-view images. Specifically, each mesh vertex is initially assigned a color from the view with maximum cosine similarity between the vertex normal and the viewing direction. This strategy ensures optimal texture quality by prioritizing views that provide the most direct observation of each surface point. Vertices at the seams between viewpoints are smoothed using Laplacian smoothing\cite{sorkine2004laplacian} to reduce color inconsistencies. Finally, we fill the colors of invisible vertices by propagating colors from visible vertices to adjacent invisible ones, producing a completely colored mesh with high-resolution details.


\vspace{-1.5mm}
\subsection{Scenario Synthesis}
\label{sec:data-synthesis}
The refined 3D assets are integrated into driving scenes through a multi-stage pipeline: 
\textbf{1) 3D Reconstruction and Object Placement.} 
The road surface is reconstructed from multi-frame aggregated LiDAR scans using \cite{huang2023nksr}. If LiDAR data is not available, a learning-based multi-view stereo~\cite{afnet} is employed for 3D geometry reconstruction. To prevent collisions between inserted 3D assets and existing road users, an object tracker estimates the trajectories of vehicles and pedestrians. Assets are placed only in collision-free regions.
\textbf{2) HDR Environment Lighting and Foreground Rendering.}
Foreground assets are rendered using a ray-casting rendering engine~\cite{blender2024}. To ensure accurate rendering of lighting and shadows, High dynamic range (HDR) environment lighting is composed by blending 6 surrounding LDR images with direct sunlight.
Depth estimation and semantic segmentation are applied to construct occlusion masks between foreground assets and static backgrounds, ensuring realistic layering.
\textbf{3) Image Composition and Post-processing.}
The rendered foreground assets are then composited with the background image or 3D models~\cite{wang2021neus,kerbl20233dgaussian} through alpha blending.
To further enhance realism, a video-based harmonizer~\cite{ke2022harmonizer} is employed to adjust the color temperature, contrast, and saturation of foreground renderings, resulting in more harmonious composition results. Visualization examples are presented in Fig.~\ref{fig:pipeline} and Fig.~\ref{fig:insert}.

\section{Experiments}
\subsection{Experimental Settings}
\textbf{Dataset.} We evaluated the performance of our proposed hybrid 3D asset generation method using two public datasets: Google Scanned Objects (GSO)~\cite{downs2022google-gso} and 3DRealCar~\cite{du20243drealcar}. The GSO dataset contains approximately 1,000 scanned objects. Following the valuation protocol of SyncDreamer~\cite{liu2023syncdreamer}, 30 selected objects from various categories were chosen as the evaluation dataset. For the 3DRealCar dataset, which contains real-world car captures, 15 vehicles with diverse appearances and shapes were selected. For each car, we chose one normal viewpoint and one difficult viewpoint to evaluate the texture quality of reconstructed meshes, resulting in the \textit{normal} and the \textit{hard} evaluation set.

We also evaluated the framework's effectiveness on two downstream tasks. For 2D detection, we use our self-collected real-driving data to evaluate performance on rare categories, specifically the Traffic A-frame sign and the warning tripod. The training set consists of 160,000 and 15,0000 2D bounding box annotations for these categories, respectively, while the test set contains 2,400 and 2,100 annotated samples. For 3D detection, we employed the NuScenes dataset, focusing on three rare categories (traffic cones, construction vehicles, and motorcycles). We use 700 street-view scenes for training and 150 for validation.

\textbf{Baselines.} For 3D asset generation, we compare our method with the following baselines, 1) Magic123~\cite{qian2023magic123}: A SDS-based 2D lifting method for single-view reconstruction. 2) SyncDreamer~\cite{liu2023syncdreamer}: A synchronized multi-view diffusion model for 3D reconstruction. 3) Wonder3D~\cite{long2024wonder3d}: A multi-view cross-domain 2D diffusion model. 4) CRM~\cite{wang2024crm}: A convolutional reconstruction model that generates multi-view images and canonical coordinate maps (CCMs). 5) InstantMesh~\cite{xu2024instantmesh}: A feed-forward framework trained with differentiable mesh rendering for high-quality 3D mesh generation. For 2D detection, we used FCOS\cite{tian2021fcos} as our baseline, and for 3D detection, BEVFormer\cite{li2024bevformer} was adopted.

\begin{table}[t]
    \begin{centering}
    \caption{\label{tab:geometry}Results on GSO dataset. \textbf{Bold}: Best. CD: Chamfer Distance.}
    \vspace{-2mm}
    \renewcommand{\arraystretch}{1.1}
    \begin{tabular}{ccccccc}
\toprule 
 Method & CD$\downarrow$ & Volume IoU$\uparrow$ & PSNR$\uparrow$ & SSIM$\uparrow$ & LPIPS$\downarrow$\\
\midrule 
Magic123 & 0.0516 & 0.4528 & 12.69 & 0.7984 & 0.2442\\
SyncDreamer  & 0.0261 & 0.5421 & 14.00 & 0.8165 & 0.2591\\
Wonder3D  & 0.0199 &  0.6244 & 13.31 & 0.8121 & 0.2554\\
CRM  & 0.0173 &  0.6286 & 16.22 & 0.8381 & 0.2143\\
InsMesh  & 0.0191 &  0.5810 & 16.84 & 0.8408 & 0.1749\\
Ours & \textbf{0.0164} &  \textbf{0.6731} & \textbf{19.05} & \textbf{0.8724} & \textbf{0.1255}\\
    \bottomrule 
    \end{tabular}
    \label{table:results-gso}
    \vspace{-2mm}
    \end{centering}
\end{table}

\begin{table}[t]
    \begin{centering}
    \caption{\label{tab:geometry}Results for Different Methods on RealCar360 \textbf{normal/hard} Evalation Set. \textbf{Bold}: Best. CD: Chamfer Distance.}
    \vspace{-2mm}
    \renewcommand{\arraystretch}{1.25}
    \begin{tabular}{@{\hspace{-3pt}}c@{\hspace{3pt}}c@{\hspace{0pt}}c@{\hspace{-1pt}}c@{\hspace{3pt}}c@{\hspace{3pt}}c}
\toprule 
 Method & CD$\downarrow$ & Volume IoU$\uparrow$ & PSNR$\uparrow$ & SSIM$\uparrow$ & LPIPS$\downarrow$\\
\midrule 
Wonder3D  & 0.0283/0.0322 &  0.48/0.40 & 14.50/13.90 & 0.768/0.793 & 0.204/0.263\\
CRM  & 0.0240/0.0330 &  0.54/0.43 & 15.29/14.42 & 0.808/0.775 & 0.167/0.180\\
InsMesh  & 0.0190/0.0256 &  0.69/0.56 & 16.53/14.76 & 0.845/0.794 & 0.157/0.177\\
Ours & \textbf{0.0170/0.0184} &  \textbf{0.76/0.70} & \textbf{16.98/16.51} & \textbf{0.857/0.851} & \textbf{0.128/0.133}\\
    \bottomrule 
    \end{tabular}
    \label{table:results-realcar360}
    \vspace{-6mm}
    \end{centering}
\end{table}

\textbf{Implementation Details.}
For 3D asset generation, the number of mesh refinement steps is set to 20, with hyperparameters for different losses set as \(\lambda_{mask} = 1\) and \(\lambda_{lap} = 0.5\). The inference time for the asset generation method is around 30 seconds on a single RTX 3090 GPU. For data synthesis, it takes approximately 20 minutes to synthesize a 2-min video (12 FPS) on a single RTX 3090 GPU.

\subsection{Evaluation of 3D Asset Generation}
\textbf{Qualitative Results.} We selected representative cases from GSO, 3DRealCar, and web image datasets for comparison, as shown in Figure~\ref{fig:results}. Our method demonstrates superior geometry quality and texture fidelity compared to baseline approaches. For example, our reconstructions capture clear clock numerals in Fig.~\ref{fig:results}(a) and precise vehicle details like rear-view mirrors and wheel contours in Fig.~\ref{fig:results}(d)(e). Even with internet-sourced data exhibiting varied lighting and geometric complexities, our method maintains high-quality reconstruction, demonstrating excellent viewpoint invariance and generalization capability. In contrast, other methods suffer from noisy geometry (CRM), lack of fine details (InstantMesh), or limited viewpoint generalization (Wonder3D). These improvements stem directly from our iterative mesh optimization and enhanced texture fusion strategy.

\textbf{Quantitative Results.} On the GSO dataset, our method outperforms all baseline approaches across geometric and visual quality metrics. Compared to the second-best method, we achieve a 14.1\% decrease in Chamfer Distance and a 7.1\% improvement in Volume IoU. For visual quality, our approach yields significant gains with a 13.1\% increase in PSNR, 3.8\% improvement in SSIM, and 28.2\% reduction in LPIPS. On the RealCar360 dataset's normal/difficult viewpoint evaluation, our method not only achieves the best results on both subsets but also exhibits minimal performance degradation on difficult viewpoints, validating its robustness in 360-degree view generalization. 

\setlength{\tabcolsep}{8pt}  
\begin{table}[th]
\vspace{-3mm}
\centering
\caption{Comparison of 2D Detection on the Self-Collected Dataset.}
\vspace{-2mm}
\label{tab:detection_results}
\begin{tabular}{lcc|ccc}
\toprule
\multirow{2}{*}{Training data} & \multicolumn{2}{c}{A-frame Sign} & \multicolumn{2}{c}{Warning Tripod} \\
\cmidrule(lr){2-3} \cmidrule(lr){4-5}
 & AP$\uparrow$(\%) &\multicolumn{1}{c}{AR$\uparrow$ (\%)}  & AP$\uparrow$ (\%) & AR$\uparrow$ (\%) \\
\midrule
real & 24.3 & 39.2 &  72.4 &  73.3 \\
real+syn data & 27.4\raisebox{-0.6ex}{\textsuperscript{\textcolor{green}{(+3.1)}}} & 43.1\raisebox{-0.6ex}{\textsuperscript{\textcolor{green}{(+3.9)}}} & 76.5\raisebox{-0.6ex}{\textsuperscript{\textcolor{green}{(+4.1)}}} & 77.1\raisebox{-0.6ex}{\textsuperscript{\textcolor{green}{(+3.8)}}} \\
\bottomrule
\label{table:2d}
\vspace{-5mm}
\end{tabular}
\end{table}

\setlength{\tabcolsep}{2.5pt}  
\begin{table}[th]
\vspace{-5mm}
\centering
\caption{Comparison of 3D Detection on NuScenes Dataset.}
\vspace{-2mm}
\begin{tabular}{l@{\hskip 5pt}cc|ccc}
\toprule
& \multicolumn{2}{c}{3D Detection} & \multicolumn{3}{c}{Per-category AP↑} \\
\cmidrule(lr){2-3} \cmidrule(lr){4-6}
Training data & NDS↑ & \multicolumn{1}{c}{mAP↑} & motorcycle & cone & vehicle \\
\midrule
real & 40.7 & 33.5 & 36.1 & 54.1 & 10.5 \\
real+20\%syn data & 42.4 \raisebox{-0.6ex}{\textsuperscript{\textcolor{green}{(+1.7)}}}
& 35.6 \raisebox{-0.6ex}{\textsuperscript{\textcolor{green}{(+2.1)}}}
& 40.1 \raisebox{-0.6ex}{\textsuperscript{\textcolor{green}{(+4.0)}}}
& 54.2 \raisebox{-0.6ex}{\textsuperscript{\textcolor{green}{(+0.1)}}}
& 12.7 \raisebox{-0.6ex}{\textsuperscript{\textcolor{green}{(+2.2)}}} \\
real+50\%syn data & 41.6 \raisebox{-0.6ex}{\textsuperscript{\textcolor{green}{(+0.9)}}}
& 35.2 \raisebox{-0.6ex}{\textsuperscript{\textcolor{green}{(+1.7)}}}
& 40.5 \raisebox{-0.6ex}{\textsuperscript{\textcolor{green}{(+4.4)}}}
& 54.7 \raisebox{-0.6ex}{\textsuperscript{\textcolor{green}{(+0.6)}}}
& 10.4 \raisebox{-0.6ex}{\textsuperscript{\textcolor{red}{(-0.1)}}} \\
\bottomrule
\end{tabular}
\label{tab:3DOD_comparison}
\vspace{-5mm}
\end{table}


\subsection{Downstream Evaluation on 2D and 3D Object Detection}
\textbf{3D Object Detection}: To evaluate the effectiveness of our synthetic data for rare object detection, we generated additional training samples and mixed them with real NuScenes data in varying proportions to train BEVFormer\cite{li2024bevformer}. As shown in Table \ref{tab:3DOD_comparison}, incorporating 20\% additional synthetic data significantly improved performance, with NDS increasing by 1.7\% and mAP by 2.1\%. The improvement was particularly pronounced for motorcycles (+4.0\%), which have limited training samples in real datasets. 
These results demonstrate the effectiveness of synthesized data for rare object detection with appropriate mixing proportions.

\textbf{2D Object Detection}:
To further evaluate the effectiveness of our synthesized data on the 2D detection task, we augmented a self-collected driving dataset by synthesizing 40,026 A-frame signs and 19,544 warning tripods, along with the corresponding 2D automated annotations. The synthesized data accounted for approximately 20\% of the total real training data. As shown in Table~\ref{table:2d}, adding synthetic data gives superior performance on rare categories, with a 3.5\% increase in average precision (AP) and a 4.0\% increase in average recall (AR). 

These results demonstrate that our high-fidelity synthetic data effectively enhances both 2D and 3D detection performance, particularly for rare and safety-critical categories in autonomous driving scenarios.

\begin{figure*}[t]
\centering
 \includegraphics[width=1.0\textwidth]{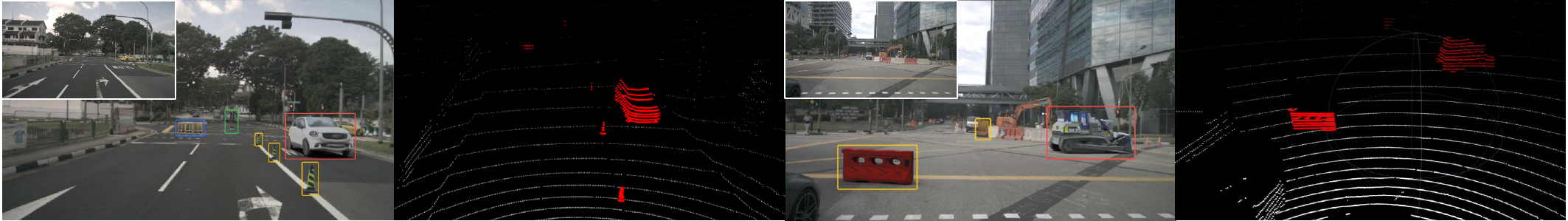}
 \begin{footnotesize}
    \begin{tabularx}
    {0.95\linewidth}{c *2{>{\centering}X}} 
    \hspace*{3.3cm}Scene A &\hspace*{7.32cm}Scene B
    \end{tabularx}
 \end{footnotesize}
\vspace{-3mm}
\caption{\label{fig:insert}Synthetic Examples with Camera and LiDAR Simulation. High-quality mined 3D assets enable straightforward synthesis of rare driving scenarios.}
\vspace{-4.5mm}
\label{fig:ablation1}
\end{figure*}

\begin{figure}[t]
\centering
 \includegraphics[width=0.4\textwidth]{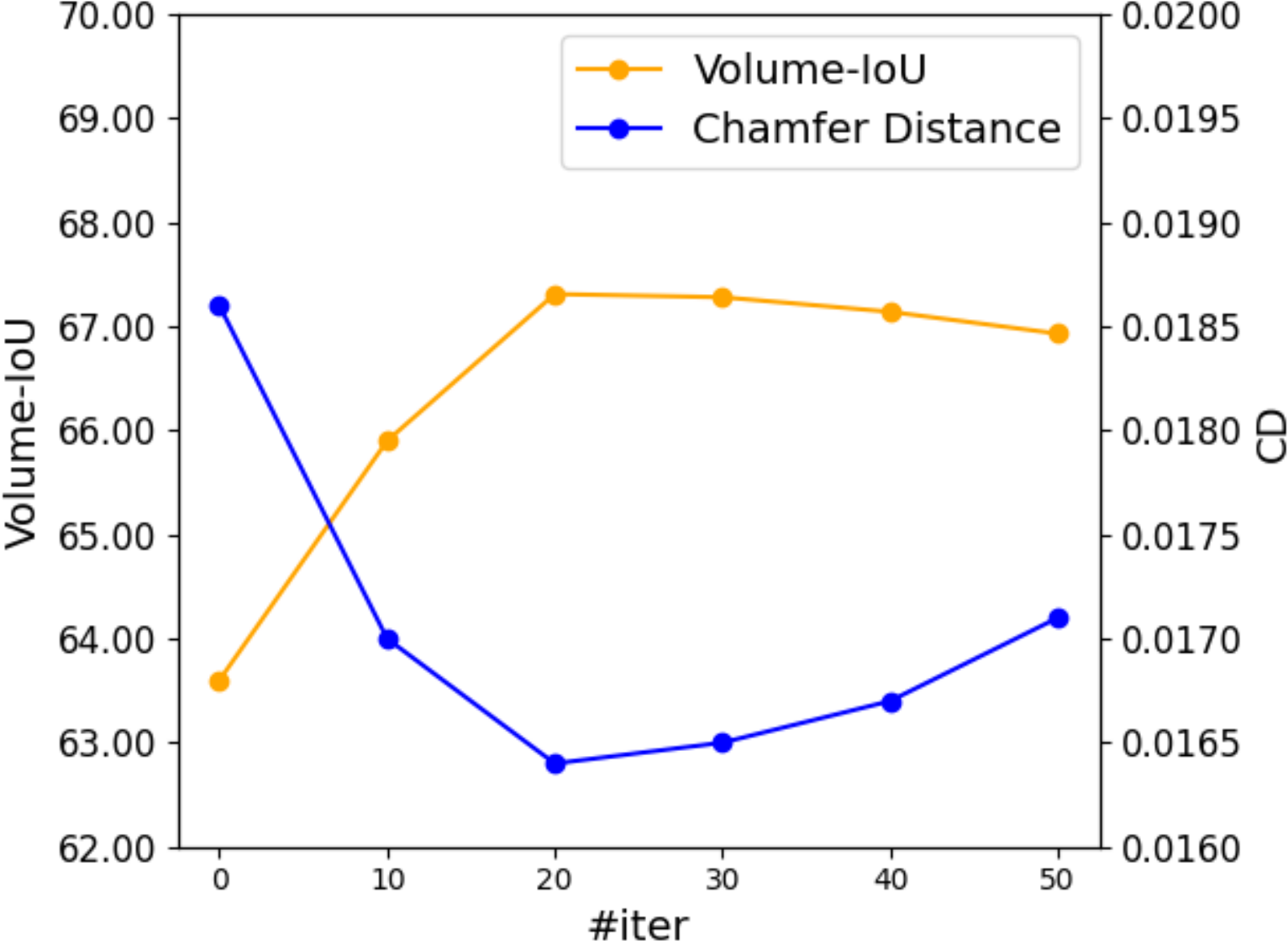}
\vspace{-4mm}
\caption{\label{fig:iter} Geometric Precision \textit{w.r.t} Optimization Iteration.}
\vspace{-1mm}
\label{fig:ablation1}
\end{figure}

\begin{figure}[t]
\centering
 \includegraphics[width=0.45\textwidth]{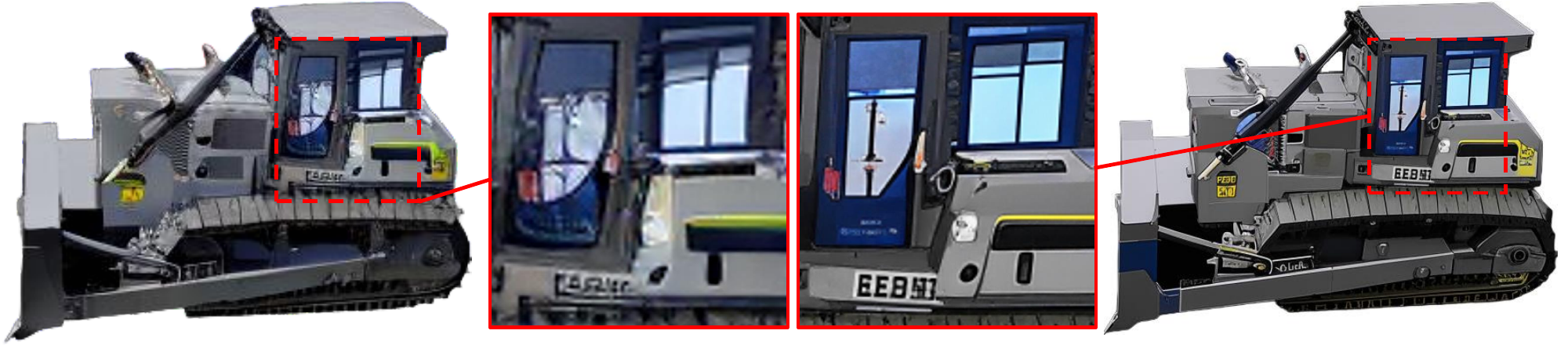}
\begin{footnotesize}
\begin{tabularx}{0.8\linewidth}{l *2{>{\raggedright\arraybackslash}X}} 
    \hspace*{0cm}w/o Texture Fusion &\hspace*{0.6cm} w/ Texture Fusion
    \end{tabularx}
    \end{footnotesize}
\vspace{-3mm}
\caption{Effect of the Proposed Texture Fusion Strategy}
\vspace{-6mm}
\label{fig:ablation0}
\end{figure}

\subsection{Ablation Study}
1) \textit{Performance w.r.t Iteration Steps of Mesh Refinement}:
As shown in Fig. \ref{fig:ablation1}, we report the F-score and Chamfer Distance at each iteration to measure the convergence of mesh refinement. It can be observed that the geometric metrics converge after around 20 iterations. With an increasing number of iterations, the number of mesh faces increases, which raises the difficulty of geometry optimization. Therefore, to balance accuracy and efficiency, we perform 20 steps for each mesh.

2) \textit{Effect of Texture Fusion}: To demonstrate the importance of texture fusion strategy, we render images of the same mesh with and without texture fusion. As shown in Fig. \ref{fig:ablation0}, the mesh with texture fusion appears clear and has sharp edges, while the mesh without texture fusion exhibits blurriness. The high resolution of the texture ensures the clearness and fidelity of synthesized data.

\subsection{Asset Bank and Scenario Synthesis}
Through our automated asset mining and generation framework, we have constructed an Asset Bank containing over 2000 high-quality 3D assets. This asset bank encompasses various non-whitelist objects encountered in autonomous driving scenarios, including road obstacles, temporary traffic facilities (portable traffic lights, movable signage), irregular vehicles, and large and small fallen objects. These assets are crucial for testing the perception robustness and capabilities of autonomous driving systems in non-typical and safety-critical scenarios. 

Figure~\ref{fig:insert} shows the results of scenario synthesis on the nuScenes dataset using our generated 3D assets. The left side presents the camera simulation results, while the right side shows the corresponding LiDAR simulation results. In Scene A, we inserted oncoming vehicles, traffic cones, temporary traffic lights, and obstacle boxes to simulate the generalizability of autonomous driving vehicles to temporary changes in the environment. In Scene B, we introduced water barriers, barricades, and bulldozers to test the perception capabilities of autonomous driving systems for irregular vehicles and obstacles. Our approach demonstrates the ability to perform realistic sensor simulation for autonomous driving across various scenarios, particularly those involving safety-critical situations.



\section{Conclusion}
In this work, we introduced a novel framework for automated asset mining and synthetic data generation designed for autonomous driving simulations. Our methodology leverages state-of-the-art 3D generation techniques, alongside improved geometry refinement and texture fusion strategies. Our pipeline enables automatic asset mining and 3D asset reconstruction from both data collected from real-world and web search engines, providing high-quality, scalable, and low-cost digital assets. Experiment results demonstrated superior scalability, robustness, and geometry accuracy over existing technologies. We hope this work will inspire the self-driving community, contributing to safer and more robust autonomous driving perception.




\bibliographystyle{IEEEtran}
\balance
\bibliography{paper}

\end{document}